\documentclass{article}

\usepackage[ruled, vlined, noend]{algorithm2e}
\usepackage{amssymb,amsmath,amsthm,accents}
\usepackage{multicol}
\usepackage{multirow}
\usepackage{graphicx} 
\usepackage{amsmath,amsfonts,bm}

\def\1{\bm{1}}

\DeclareMathAlphabet{\mathsfit}{\encodingdefault}{\sfdefault}{m}{sl}
\SetMathAlphabet{\mathsfit}{bold}{\encodingdefault}{\sfdefault}{bx}{n}

\usepackage{accents}
\newlength{\dhatheight}

\usepackage[utf8]{inputenc} % allow utf-8 input
\usepackage[T1]{fontenc}    % use 8-bit T1 fonts
\usepackage{hyperref}       % hyperlinks
\usepackage{url}            % simple URL typesetting
\usepackage{booktabs}       % professional-quality tables
\usepackage{amsfonts}       % blackboard math symbols
\usepackage{nicefrac}       % compact symbols for 1/2, etc.
\usepackage{microtype}      % microtypography
\usepackage{xcolor}         % colors

\usepackage[accepted]{icml2024}

\icmltitlerunning{Interpreting Microbiome Relative Abundance Data Using Symbolic Regression 
}

\usepackage{tcolorbox}

\begin{document}

\twocolumn[
\icmltitle{Interpreting Microbiome Relative Abundance Data Using Symbolic Regression 
% and Knowledge Distillation
}



\icmlsetsymbol{equal}{*}

\begin{icmlauthorlist}
\icmlauthor{Swagatam Haldar}{tue}
\icmlauthor{Christoph Stein-Thoeringer}{tue,M3,cluster}
\icmlauthor{Vadim Borisov}{tue,M3,cluster}

\end{icmlauthorlist}

\icmlaffiliation{tue}{University of Tübingen, Germany}
\icmlaffiliation{M3}{University Clinic Tuebingen, Germany }
\icmlaffiliation{cluster}{Cluster of Excellence "Controlling Microbes to Fight Infections", University of Tübingen, Germany}

% \affil[3]{\orgdiv{Department of Internal Medicine I and M3 Research Institute}, \orgname{University Clinic Tuebingen}, \orgaddress{\street{Otfried-Mueller-Strasse 37}, \city{Tuebingen}, \postcode{72076}, \country{Germany}}}

% \affil[4]{\orgdiv{Cluster of Excellence "Controlling Microbes to Fight Infections"}, \orgname{University of Tuebingen}, \city{Tuebingen}, \postcode{72076}, \country{Germany}}


\icmlcorrespondingauthor{Swagatam Haldar}{swagatam.haldar9@gmail.com}

\icmlkeywords{Explainable AI, Symbolic Regression, Microbiome data}

\vskip 0.3in
]



\printAffiliationsAndNotice{\icmlEqualContribution} % otherwise use the standard text.

\begin{abstract}
Understanding the complex interactions within the microbiome is crucial for developing effective diagnostic and therapeutic strategies. Traditional machine learning models often lack interpretability, which is essential for clinical and biological insights. This paper explores the application of symbolic regression (SR) to microbiome relative abundance data, with a focus on colorectal cancer (CRC). SR, known for its high interpretability, is compared against traditional machine learning models, e.g., random forest, gradient boosting decision trees. These models are evaluated based on performance metrics such as F1 score and accuracy. We utilize 71 studies encompassing from various cohorts over 10,000 samples across 749 species features. Our results indicate that SR not only competes reasonably well in terms of predictive performance, but also excels in model interpretability. SR provides explicit mathematical expressions that offer insights into the biological relationships within the microbiome, a crucial advantage for clinical and biological interpretation. Our experiments also show that SR can help understand complex models like XGBoost via knowledge distillation.
To aid in reproducibility and further research, we have made the code openly available at \url{https://github.com/swag2198/microbiome-symbolic-regression}.
%This approach positions SR as a powerful tool for achieving both high predictive accuracy and enhanced transparency in microbiome research.
\end{abstract}

\section{Introduction}

The microbiome, a complex and diverse ecosystem located across various body sites, plays a crucial role in host functions impacting health and disease \cite{curry2021takes, gomaa2020human, durack2019gut}. This study focuses on the relative abundances of stool microbial species in both healthy individuals and patients with colorectal cancer (CRC), although our methods are applicable to other compositional data as well. Understanding the key microbial players in CRC is essential for developing targeted diagnostic and therapeutic strategies.

Recent advancements in machine learning (ML) have significantly improved the analysis of complex biological data. Traditional ML approaches, such as linear models, decision trees, ensemble methods, and neural networks, have been widely used for this purpose \cite{gordon2022data, hernandez2022machine}. However, these methods often do not provide explainable insights, which are crucial for clinical and biological interpretations.

Symbolic regression (SR) emerges as a powerful alternative, offering clear, interpretable mathematical models of data relationships. Unlike standard ML models that often act as \emph{blackboxes}, SR provides explicit equations that are inherently interpretable. In our study, we leverage SR to dissect the relationships within microbiome data, implementing it through a commonly used package \texttt{gplearn}\footnote{https://github.com/trevorstephens/gplearn}, enhanced by custom functions suitable for microbiome analyses. This helps to construct explanatory graphs that detail microbial interactions and their implications in CRC.

The contributions of this paper are twofold:
\begin{itemize}
    \item Introduction of a novel application of symbolic regression to microbiome data, demonstrating its superior explainability properties compared to traditional ML approaches.
    \item Development and validation of a custom set of functions that extends the capability of SR tools for enhanced biological data interpretation, enabling more effective communication of complex microbial relationships to non-specialist audiences.
\end{itemize}

% Symbolic regression, as used in this study, addresses these challenges by deriving mathematical models that not only predict outcomes with high accuracy but also give explicit formulas that describe the relationships within the data. This approach is highly beneficial for elucidating the biological mechanisms at play, making the results both scientifically insightful and practically applicable in medical and ecological contexts.

\section{Background}

\subsection{Background on Microbiome Relative Abundance Data}

Microbiome relative abundance data quantifies the proportion of various microbial entities within a sample. Unlike absolute abundance, which counts the exact number of organisms, relative abundance indicates the relative percentage each microorganism contributes to the total sample \cite{neu2021defining}. 
%The gut microbiome comprises a complex array of bacteria, viruses, fungi, and other microorganisms crucial for human health, influencing digestion, immune function, and disease resistance \cite{cullin2021microbiome}.
The inherent complexity of microbiome relative abundance data represents significant analytical challenges. Each data point represents a percentage, creating values ranging from 0 to 100 after normalization. The dataset is characterized by high dimensionality, with up to 2000 features that are mostly sparse (containing many zeros), and each row sums to 100. This necessitates specialized analytical techniques capable of handling data sparsity and high dimensionality.

Variations in the gut microbiome composition have been linked to a wide range of health conditions, from psychological disorders to diabetes and obesity, underscoring the potential of microbiome studies to impact various medical fields \cite{gomaa2020human, cullin2021microbiome}.

\subsection{Background on Symbolic Regression}
Supervised machine learning tries to understand the relationship between some input variables (e.g., features of patients) and output variable(s) or target(s) (e.g., presence of a disease) by updating the parameters of a model using past observations.
Many machine learning models, such as neural networks, decision trees, and others, can achieve high accuracy but often lack transparency. Symbolic regression (\textbf{SR}) is a technique that learns the input-target relationship using \emph{transparent} equations involving mathematical operators. 
The operator set can be simple arithmetic operations ($+$, $-$, $\times$, $/$), functions of arbitrary arity (e.g., $\log(.)$, $\max(.,.)$) including the flexibility of including custom, domain specific functions.
SR has been applied to noisy scientific data for equation discovery~\cite{schmidt2009SReureqa, bongard2007automatedSReureqa, Eureqa_SRforEquations}, and recently used as a supervised classification method~\cite{korns2018gpsr_classification,sredt_Fong_Motani_2024}.

SR learns the best mathematical expression from data using genetic programming or evolutionary algorithms which are popularized in the 90s~\cite{koza1994geneticprogramming} and even modern SR algorithms incorporate GP as their primary workhorse~\cite{petersen2021deepSR,cranmer2020pysr,Stephens}. The core idea of GP is to maintain a \emph{population} of equations and evaluate their \emph{fitness} (e.g., for classification it could be accuracy or any other metric), and subsequently perform \emph{mutation} and \emph{selection} of best-performing equation from the population.

\section{Symbolic Regression for Relative Abundance Data}

% We can keep or discard this section later based on if we come up with some novel method later on ....

In this section, we discuss the modifications we have made on top of off-the-shelf symbolic regression to 
% make it more amenable to relative abundance data.
interpret relative abundance data better.
Since we primarily care about the presence (or absence) of one bacterium (or multiple species together), we introduce a few custom functions that enforce such relationships:

\begin{equation}
\begin{aligned}
\text{presence}(x) &= 
\begin{cases} 
1, & \text{if } x > 0, \\
0, & \text{otherwise}.
\end{cases}
\end{aligned}
\end{equation}

\vspace{-.4cm}

\begin{equation}
\begin{aligned}
\text{presence\_both}(x_1, x_2) &= 
\begin{cases} 
1, & \text{if } x_1 > 0 \text{ and } x_2 > 0, \\
0, & \text{otherwise}.
\end{cases}
\end{aligned}
\end{equation}

Similarly, \textsf{absence} and \textsf{absence\_both} are defined to indicate the absence of a single species or two species together. We also define:

\begin{equation}
\begin{aligned}
\text{ifelse}(x_1, x_2) &= 
\begin{cases} 
x_1, & \text{if } x_1 > x_2, \\
x_2, & \text{otherwise}.
\end{cases}
\end{aligned}
\end{equation}

% \vspace{-.1cm}
to take a bacterium having higher abundance. We denote this symbolic regression model with these custom functions as \textbf{SRf} and evaluate its performance along with other baselines.

\section{Experiments}
\label{sec_experiments}

The primary goals of our experiments is to first verify the accuracy of SR against other classification models, and second, to establish the effectiveness of SR in identifying the most influential gut microbial species for predicting CRC.

\noindent\textbf{Datasets:} This study utilized $71$ datasets from the \texttt{curatedMetagenomicData} library \cite{datasets}. From these datasets, we selected $11,137$ samples classified as either \textit{healthy} or \textit{CRC}. This selection resulted in a highly imbalanced dataset, with $10,473$ healthy individuals and only $664$ CRC patients. To address this imbalance, we performed random undersampling of the healthy individuals. Detailed statistics of the datasets and our data processing steps are provided in Appendix~\ref{ap_datasets}.

\noindent\textbf{ML models:} We selected a set of the most commonly used ML models of varying degrees of complexity and interpretability \cite{borisov2022deep}: Logistic Regression (\textbf{LR}), Decision Tree (\textbf{DT}), Random Forest (\textbf{RF}) \cite{breiman2001random},  and XGBoost (\textbf{XG}) \cite{chen2016xgboost} for benchmarking against the Symbolic Regression (\textbf{SR}) method. We abbreviate Symbolic Regression with our custom functions added as \textbf{SRf}. We used the scikit-learn~\cite{scikit-learn} and gplearn~\cite{Stephens} libraries for ML models and our implementations of \textbf{SRf}.

We investigated the following research questions:

\noindent\textbf{RQ1:} How does SR compare with other ML models in terms of classification accuracy and F1 score? What is the effect of incorporating custom functions?

\noindent\textbf{RQ2:} How can we identify the influential microbial species for CRC prediction from the expressions provided by SR?
% , what are some potential issues with it (e.g., high variance in expressions across runs)? \mc{may be geographical segregation comes here/need to highlight how SR helps in accelerated scientific discovery}

\noindent\textbf{RQ3:} Can SR be used as a \emph{knowledge distillation} method to understand or explain a complex, black-box model (e.g., XGBoost) with high accuracy but low interpretability?
%How does it compare to using SR directly to fit the data?

\begin{table}[tb]
\small
\centering
\caption{Metrics (balanced) for healthy vs. CRC binary classification problem. Result shows SR, although below par compared to complex models like RF \& XG (bolded), performs similarly (with lower variance) to other interpretable models like LR and DT.}
\vspace{.2cm}
\begin{tabular}{lcc}
\toprule
\textbf{Models}     & \textbf{Accuracy ($\uparrow$)}  &  \textbf{F1 score ($\uparrow$)} \\
\midrule
LR & $0.75$ \tiny{$\pm0.023$} & $0.72$  \tiny{$\pm 0.023$} \\
DT & $0.74$ \tiny{$\pm 0.024$} & $0.70$  \tiny{$\pm 0.031$} \\
RF & $0.82$ \tiny{$\pm 0.015$} &  $0.78$  \tiny{$\pm 0.018$} \\
XG & $\mathbf{0.83}$ \tiny{$\pm 0.015$} &  $\mathbf{0.80}$  \tiny{$\pm 0.015$} \\
SR & $0.75$ \tiny{$\pm 0.011$} &          $0.70$  \tiny{$\pm 0.011$} \\
SRf & $0.75$ \tiny{$\pm 0.015$} &          $0.69$  \tiny{$\pm 0.017$} \\
\bottomrule
\end{tabular}
\label{tab:benchmarks}
\end{table}

\begin{figure}[ht]
    \centering
    % \vspace{-0.45cm}
    \includegraphics[width=\columnwidth]{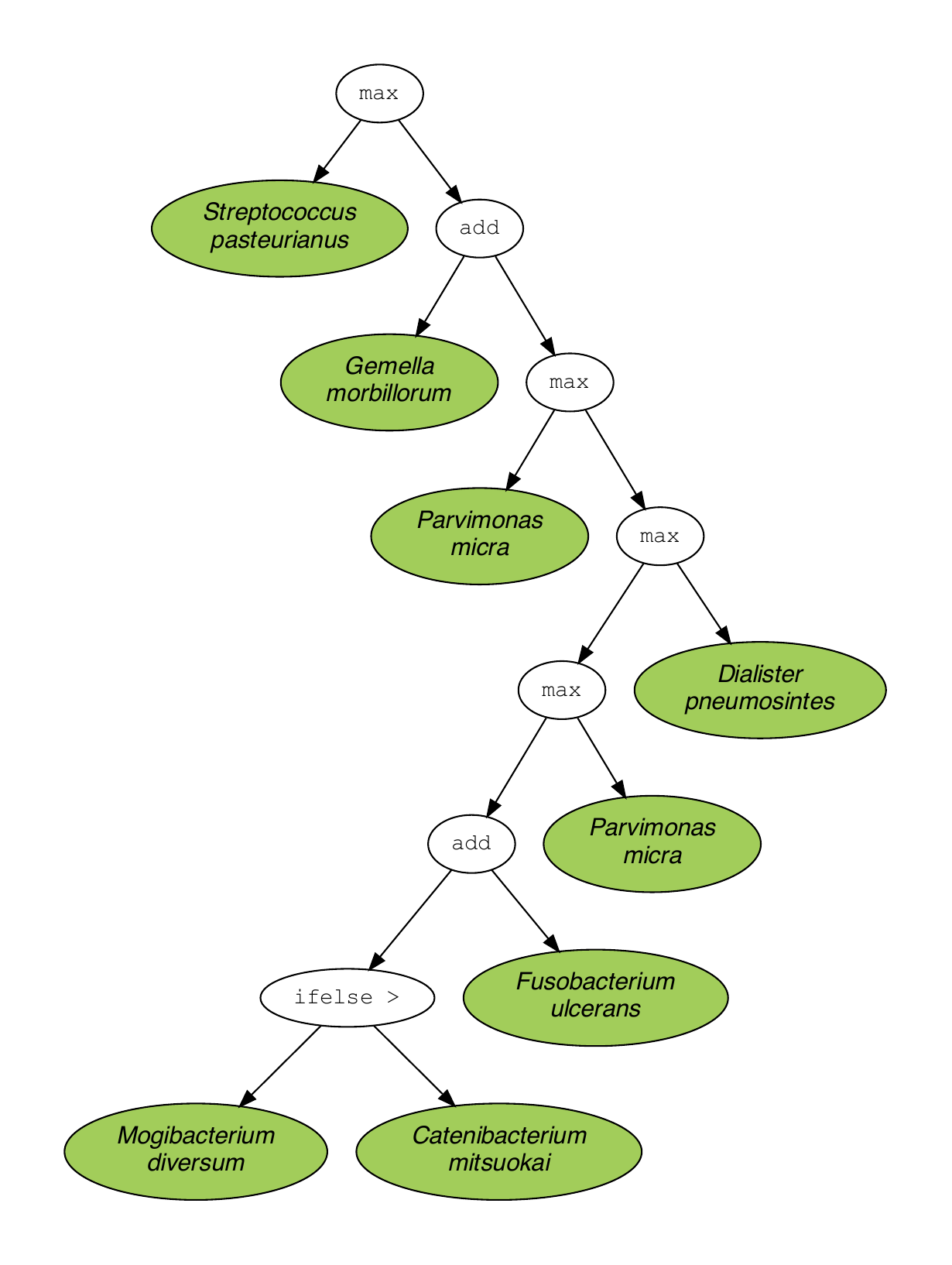}
    \caption{\small{Visualization of an SRf expression learned using \texttt{gplearn}. The added custom function \textsf{ifelse} helped to reduce its size down to 15 nodes while the SR (without custom functions) expression for the same run had 25 nodes (figure in Appendix~\ref{appdx:sr-vs-srf}).}
    }
    \label{fig:srfmodel}
    % \vspace{-1.5cm}
\end{figure}

\subsection{Benchmarking}
\label{sec:bench}
We report the classification accuracy and F1 score of selected ML models in classifying the test set patients into \textit{healthy} or \textit{CRC} categories. As mentioned earlier (Sec.~\ref{sec_experiments}), we have drawn $20$ random (balanced) subsamples of our data and provide the mean and standard deviation of the metrics in Table~\ref{tab:benchmarks}. We observed that although the SR models perform below par compared to complex but uninterpretable models like XG and RF, they performed similarly to other interpretable models (LR and DT), and notably with lower variance.

Interestingly, between SR and SRf, we note a difference in their total expression lengths (number of nodes in a graph). The mean length of SR is $19.55$, while that for SRf is $14.55$. In $16$ out of the $20$ runs, the expressions found by SRf are smaller than those of SR. Thus, adding custom functions (such as \textsf{presence}, \textsf{absence}, \textsf{ifelse} etc.) has the benefit of reducing the learned expression size in our experiments. We show one SRf expression tree learned in Figure~\ref{fig:srfmodel} (corresponding SR expression in Appendix~\ref{appdx:sr-vs-srf}).

We emphasize here that even though RF and XG models are composed of many decision trees, they are not considered inherently interpretable models. SR and SRf are advantageous over them as they identify few variables and show explicit mathematical dependencies between them as shown in Figure~\ref{fig:srfmodel}. We also investigate how SR can be used as an interpretable surrogate for complex models in Section~\ref{sec:kd}.

\subsection{Discovering species from SR expressions}
\label{sec:discovery}
Having established that SR performs comparably well for this task, we now discuss how the interpretable expressions can be leveraged to discover influential bacteria associated with CRC (or healthy) individuals. In Table~\ref{tab:bacteria}, we list the bacteria species that appear most frequently in the $10$ SRf expressions and discuss their significance, if any.

Validating this feature importance analysis is challenging due to several factors: absence of ground truth, expensive wet lab validation, and variability in dataset sources. However, multiple independent studies have shown that bacteria from our top $10$ list are related to the development or progression of CRC cancer: \textit{Gemella morbillorum}~\cite{shimomura2023mediation, chang2021metagenomic}, \textit{Mogibacterium diversum}~\cite{chen2024characterization}, \textit{Fusobacterium ulcerans}~\cite{cullin2021microbiome, amitay2017fusobacterium}, \textit{Parvimonas micra}~\cite{chang2021metagenomic, zhao2022parvimonas, lowenmark2020parvimonas, osman2021parvimonas, conde2024parvimonas}, \textit{Fusobacterium nucleatum}~\cite{cullin2021microbiome, amitay2017fusobacterium, osman2021parvimonas}.

To provide a quantitative perspective, we present the mean and standard deviation values of the top 10 features in our datasets in the Appendix (Fig.~\ref{fig:mean_std}). These statistics offer insight into the variability and central tendency of the key features identified by SR.

Overall, this shows that the SRf approach can be a valuable tool for identifying key bacterial species associated with CRC, providing a foundation for further biological validation and potential therapeutic targets. 
%Future work should focus on improving the robustness of these findings through larger datasets and enhanced validation methods.

\begin{table}[tb]
\small
\centering
\caption{Top $10$ bacteria species out of 749 extracted from symbolic regression (SRf) expressions. This ordered list is computed by parsing the $20$ SRf expressions fitted on random subsamples of the data and counting how many times each bacterium appears across all the expressions.
}
\vspace{.2cm}
\begin{tabular}{lcc}
\toprule
\textbf{Bacteria}     & \textbf{Count} \\
\midrule
\emph{Gemella morbillorum} & $27$ \\
\emph{Mogibacterium diversum} & $25$ \\
\emph{Fusobacterium ulcerans} & $25$ \\
\emph{Parvimonas micra} & $25$ \\
\emph{Fusobacterium nucleatum} & $14$ \\
\emph{Dialister pneumosintes} & $10$ \\
\emph{Catenibacterium mitsuokai} & $4$ \\
\emph{Butyricicoccus pullicaecorum} & $4$ \\
\emph{Streptococcus cristatus} & $4$ \\
\emph{Romboutsia ilealis} & $3$ \\
% \emph{Clostridium sp CAG 167} & $2$ \\
% \emph{Solobacterium moorei} & $2$ \\
% \emph{Peptostreptococcus stomatis} & $2$ \\
% \emph{Streptococcus pasteurianus} & $1$ \\
% \emph{Enorma massiliensis} & $1$ \\
% \emph{Streptococcus downei} & $1$ \\
% \emph{Prevotella nigrescens} & $1$ \\
% \emph{Gemella haemolysans} & $1$ \\
% \emph{Lactobacillus crispatus} & $1$ \\
% \emph{Bacteroidales bacterium KA00251} & 1 \\
\bottomrule
\end{tabular}
\label{tab:bacteria}
\end{table}

\subsection{Knowledge Distillation from XGBoost using SR}
\label{sec:kd}
In Section~\ref{sec:bench}, we have shown that the gradient-boosting decision tree model (XG) is the best-performing model for the selected binary classification task. However, it is a black-box model consisting of $50$ decision trees, and it is not straightforward to understand how XGBoost decides the label for a particular sample. We now demonstrate how we can obtain simple expressions that achieve similar performance to XG using SRf.

\begin{figure}[ht]
    \centering
    \includegraphics[width=\columnwidth]{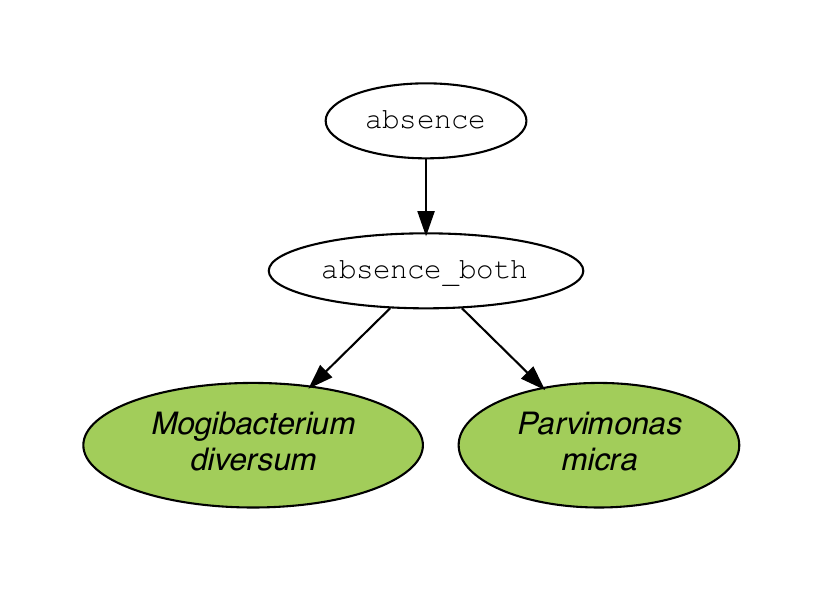}
    \vspace{-.55cm}
    \caption{Visualization of the SRf expression obtained by fitting it on predictions by XG. The expression is a composition of \textsf{absence} \& \textsf{absence\_both} functions which ultimately implies the presence of atleast one of the two identified bacteria in CRC patients. \emph{This simple expression gives the same prediction as XG (with $50$ trees) to over $80$\% of cases while being straightforward to understand.}}
    \label{fig:kd-xgboost}
    % \vspace{-.25cm}
\end{figure}

To distill the knowledge from the fitted XG model, we collect its predicted labels on the training examples, and fit the SRf models on these new labels (instead of original labels as used earlier in Sec.~\ref{sec:bench}). So now the learned symbolic expression would mimic the XG model in its predictions, thereby giving an interpretable \emph{surrogate} for the blackbox. We present the distilled expression in Figure~\ref{fig:kd-xgboost} (details in caption). Interestingly, it also features two of the most harmful bacteria as identified in the last section.

\vspace{-.2cm}
\section{Related Works}

Understanding the interpretability of machine learning models, especially in the context of microbiome research, has gained significant attention in recent years. Traditional models like logistic regression, decision trees, random forest, and XGBoost have been widely utilized for predictive tasks, including those involving microbiome data \cite{gordon2022data, hernandez2022machine}. There are tools such as TAMPA~\cite{TAMPA_Sarwal2022} that provide visualization of
metagenomics-based taxon abundance profiles. However, these models often fall short in providing clear, interpretable insights crucial for clinical and biological applications. Explainability techniques, such as LIME~\cite{ribeiro2016should} and SHAP~\cite{lundberg2017unified}, have been employed to enhance interpretability, though they still rely on post-hoc explanations.

Symbolic regression (SR) offers an alternative by generating explicit 
% mathematical 
expressions that describe the relationships within the data. Previous studies have demonstrated SR's potential for high interpretability \cite{cranmer2020discovering, kamienny2022end}. In microbiome data, SR provides direct insights into associations between microbial species and health outcomes. Additionally, the endoR framework \cite{ruaud2022endoR} interprets tree-based models by converting them into simpler, rule-based representations. Integrating SR and knowledge distillation presents a promising direction for enhancing the interpretability of machine learning models in microbiome research, facilitating a nuanced understanding of microbial interactions and their implications for health and disease.

\vspace{-.2cm}
\section{Conclusion and Future Work}

In this paper, we presented the application of the Symbolic Regression (SR) model for interpreting microbiome relative abundance data by comparing it to common machine learning models, evaluating the expressions given by SR, and showing how one can improve the compactness of the SR model by utilizing domain specific functions. In future work, we will focus on improving the robustness of these findings through larger datasets and enhanced validation methods. We also leave methods to take advantage of the sparsity present in the dataset to future work.

Overall, our study demonstrates the potential of Symbolic Regression in microbiome research, balancing predictive performance and interpretability. By offering explicit mathematical expressions, SR enables deeper understanding of biological relationships within the microbiome, crucial for clinical and biological interpretation. Additionally, knowledge distillation shows promise in simplifying complex models like XGBoost, making them more accessible for scientific inquiry and practical applications.

\bibliography{bib}

\begin{thebibliography}{105}
\providecommand{\natexlab}[1]{#1}
\providecommand{\url}[1]{\texttt{#1}}
\expandafter\ifx\csname urlstyle\endcsname\relax
  \providecommand{\doi}[1]{doi: #1}\else
  \providecommand{\doi}{doi: \begingroup \urlstyle{rm}\Url}\fi

\bibitem[hum(2012)]{human2012structure}
Structure, function and diversity of the healthy human microbiome.
\newblock \emph{nature}, 486\penalty0 (7402):\penalty0 207--214, 2012.

\bibitem[Amitay et~al.(2017)Amitay, Werner, Vital, Pieper, H{\"o}fler, Gierse, Butt, Balavarca, Cuk, and Brenner]{amitay2017fusobacterium}
Amitay, E.~L., Werner, S., Vital, M., Pieper, D.~H., H{\"o}fler, D., Gierse, I.-J., Butt, J., Balavarca, Y., Cuk, K., and Brenner, H.
\newblock Fusobacterium and colorectal cancer: causal factor or passenger? results from a large colorectal cancer screening study.
\newblock \emph{Carcinogenesis}, 38\penalty0 (8):\penalty0 781--788, 2017.

\bibitem[Asnicar et~al.(2017)Asnicar, Manara, Zolfo, Truong, Scholz, Armanini, Ferretti, Gorfer, Pedrotti, Tett, et~al.]{asnicar2017studying}
Asnicar, F., Manara, S., Zolfo, M., Truong, D.~T., Scholz, M., Armanini, F., Ferretti, P., Gorfer, V., Pedrotti, A., Tett, A., et~al.
\newblock Studying vertical microbiome transmission from mothers to infants by strain-level metagenomic profiling.
\newblock \emph{MSystems}, 2\penalty0 (1):\penalty0 e00164--16, 2017.

\bibitem[Asnicar et~al.(2021)Asnicar, Berry, Valdes, Nguyen, Piccinno, Drew, Leeming, Gibson, Le~Roy, Khatib, et~al.]{asnicar2021microbiome}
Asnicar, F., Berry, S.~E., Valdes, A.~M., Nguyen, L.~H., Piccinno, G., Drew, D.~A., Leeming, E., Gibson, R., Le~Roy, C., Khatib, H.~A., et~al.
\newblock Microbiome connections with host metabolism and habitual diet from 1,098 deeply phenotyped individuals.
\newblock \emph{Nature medicine}, 27\penalty0 (2):\penalty0 321--332, 2021.

\bibitem[Bedarf et~al.(2017)Bedarf, Hildebrand, Coelho, Sunagawa, Bahram, Goeser, Bork, and W{\"u}llner]{bedarf2017functional}
Bedarf, J.~R., Hildebrand, F., Coelho, L.~P., Sunagawa, S., Bahram, M., Goeser, F., Bork, P., and W{\"u}llner, U.
\newblock Functional implications of microbial and viral gut metagenome changes in early stage l-dopa-na{\"\i}ve parkinson’s disease patients.
\newblock \emph{Genome medicine}, 9:\penalty0 1--13, 2017.

\bibitem[Bengtsson-Palme et~al.(2015)Bengtsson-Palme, Hartmann, Eriksson, Pal, Thorell, Larsson, and Nilsson]{bengtsson2015metaxa2}
Bengtsson-Palme, J., Hartmann, M., Eriksson, K.~M., Pal, C., Thorell, K., Larsson, D. G.~J., and Nilsson, R.~H.
\newblock Metaxa2: improved identification and taxonomic classification of small and large subunit rrna in metagenomic data.
\newblock \emph{Molecular ecology resources}, 15\penalty0 (6):\penalty0 1403--1414, 2015.

\bibitem[Bongard \& Lipson(2007)Bongard and Lipson]{bongard2007automatedSReureqa}
Bongard, J. and Lipson, H.
\newblock Automated reverse engineering of nonlinear dynamical systems.
\newblock \emph{Proceedings of the National Academy of Sciences}, 104\penalty0 (24):\penalty0 9943--9948, 2007.

\bibitem[Borisov et~al.(2022)Borisov, Leemann, Se{\ss}ler, Haug, Pawelczyk, and Kasneci]{borisov2022deep}
Borisov, V., Leemann, T., Se{\ss}ler, K., Haug, J., Pawelczyk, M., and Kasneci, G.
\newblock Deep neural networks and tabular data: A survey.
\newblock \emph{IEEE Transactions on Neural Networks and Learning Systems}, 2022.

\bibitem[Breiman(2001)]{breiman2001random}
Breiman, L.
\newblock Random forests.
\newblock \emph{Machine learning}, 45:\penalty0 5--32, 2001.

\bibitem[Brito et~al.(2016)Brito, Yilmaz, Huang, Xu, Jupiter, Jenkins, Naisilisili, Tamminen, Smillie, Wortman, et~al.]{brito2016mobile}
Brito, I.~L., Yilmaz, S., Huang, K., Xu, L., Jupiter, S.~D., Jenkins, A.~P., Naisilisili, W., Tamminen, M., Smillie, C.~S., Wortman, J.~R., et~al.
\newblock Mobile genes in the human microbiome are structured from global to individual scales.
\newblock \emph{Nature}, 535\penalty0 (7612):\penalty0 435--439, 2016.

\bibitem[Chang et~al.(2021)Chang, Mishra, Cen, Tang, Ma, Wasti, Wang, Ou, Chen, and Zhang]{chang2021metagenomic}
Chang, H., Mishra, R., Cen, C., Tang, Y., Ma, C., Wasti, S., Wang, Y., Ou, Q., Chen, K., and Zhang, J.
\newblock Metagenomic analyses expand bacterial and functional profiling biomarkers for colorectal cancer in a hainan cohort, china.
\newblock \emph{Current Microbiology}, 78:\penalty0 705--712, 2021.

\bibitem[Chen et~al.(2024)Chen, Huang, Zhang, Jiang, Zeng, Li, Su, Chen, Lin, Li, et~al.]{chen2024characterization}
Chen, Q., Huang, X., Zhang, H., Jiang, X., Zeng, X., Li, W., Su, H., Chen, Y., Lin, F., Li, M., et~al.
\newblock Characterization of tongue coating microbiome from patients with colorectal cancer.
\newblock \emph{Journal of Oral Microbiology}, 16\penalty0 (1):\penalty0 2344278, 2024.

\bibitem[Chen \& Guestrin(2016)Chen and Guestrin]{chen2016xgboost}
Chen, T. and Guestrin, C.
\newblock Xgboost: A scalable tree boosting system.
\newblock In \emph{Proceedings of the 22nd acm sigkdd international conference on knowledge discovery and data mining}, pp.\  785--794, 2016.

\bibitem[Chu et~al.(2017)Chu, Ma, Prince, Antony, Seferovic, and Aagaard]{chu2017maturation}
Chu, D.~M., Ma, J., Prince, A.~L., Antony, K.~M., Seferovic, M.~D., and Aagaard, K.~M.
\newblock Maturation of the infant microbiome community structure and function across multiple body sites and in relation to mode of delivery.
\newblock \emph{Nature medicine}, 23\penalty0 (3):\penalty0 314--326, 2017.

\bibitem[Conde-P{\'e}rez et~al.(2024)Conde-P{\'e}rez, Buetas, Aja-Macaya, Martin-De~Arribas, Iglesias-Corr{\'a}s, Trigo-Tasende, Nasser-Ali, Est{\'e}vez, Rumbo-Feal, Otero-Al{\'e}n, et~al.]{conde2024parvimonas}
Conde-P{\'e}rez, K., Buetas, E., Aja-Macaya, P., Martin-De~Arribas, E., Iglesias-Corr{\'a}s, I., Trigo-Tasende, N., Nasser-Ali, M., Est{\'e}vez, L.~S., Rumbo-Feal, S., Otero-Al{\'e}n, B., et~al.
\newblock Parvimonas micra can translocate from the subgingival sulcus of the human oral cavity to colorectal adenocarcinoma.
\newblock \emph{Molecular Oncology}, 18\penalty0 (5):\penalty0 1143--1173, 2024.

\bibitem[Costea et~al.(2017)Costea, Zeller, Sunagawa, Pelletier, Alberti, Levenez, Tramontano, Driessen, Hercog, Jung, et~al.]{costea2017towards}
Costea, P.~I., Zeller, G., Sunagawa, S., Pelletier, E., Alberti, A., Levenez, F., Tramontano, M., Driessen, M., Hercog, R., Jung, F.-E., et~al.
\newblock Towards standards for human fecal sample processing in metagenomic studies.
\newblock \emph{Nature biotechnology}, 35\penalty0 (11):\penalty0 1069--1076, 2017.

\bibitem[Cranmer(2020)]{cranmer2020pysr}
Cranmer, M.
\newblock Pysr: Fast \& parallelized symbolic regression in python/julia.
\newblock \emph{Zenodo, September}, 2020.

\bibitem[Cranmer et~al.(2020)Cranmer, Sanchez~Gonzalez, Battaglia, Xu, Cranmer, Spergel, and Ho]{cranmer2020discovering}
Cranmer, M., Sanchez~Gonzalez, A., Battaglia, P., Xu, R., Cranmer, K., Spergel, D., and Ho, S.
\newblock Discovering symbolic models from deep learning with inductive biases.
\newblock \emph{Advances in neural information processing systems}, 33:\penalty0 17429--17442, 2020.

\bibitem[Cullin et~al.(2021)Cullin, Antunes, Straussman, Stein-Thoeringer, and Elinav]{cullin2021microbiome}
Cullin, N., Antunes, C.~A., Straussman, R., Stein-Thoeringer, C.~K., and Elinav, E.
\newblock Microbiome and cancer.
\newblock \emph{Cancer Cell}, 39\penalty0 (10):\penalty0 1317--1341, 2021.

\bibitem[Curry et~al.(2021)Curry, Nute, and Treangen]{curry2021takes}
Curry, K.~D., Nute, M.~G., and Treangen, T.~J.
\newblock It takes guts to learn: machine learning techniques for disease detection from the gut microbiome.
\newblock \emph{Emerging Topics in Life Sciences}, 5\penalty0 (6):\penalty0 815--827, 2021.

\bibitem[David et~al.(2015)David, Weil, Ryan, Calderwood, Harris, Chowdhury, Begum, Qadri, LaRocque, and Turnbaugh]{david2015gut}
David, L.~A., Weil, A., Ryan, E.~T., Calderwood, S.~B., Harris, J.~B., Chowdhury, F., Begum, Y., Qadri, F., LaRocque, R.~C., and Turnbaugh, P.~J.
\newblock Gut microbial succession follows acute secretory diarrhea in humans.
\newblock \emph{MBio}, 6\penalty0 (3):\penalty0 10--1128, 2015.

\bibitem[De~Filippis et~al.(2019)De~Filippis, Pasolli, Tett, Tarallo, Naccarati, De~Angelis, Neviani, Cocolin, Gobbetti, Segata, et~al.]{de2019distinct}
De~Filippis, F., Pasolli, E., Tett, A., Tarallo, S., Naccarati, A., De~Angelis, M., Neviani, E., Cocolin, L., Gobbetti, M., Segata, N., et~al.
\newblock Distinct genetic and functional traits of human intestinal prevotella copri strains are associated with different habitual diets.
\newblock \emph{Cell host \& microbe}, 25\penalty0 (3):\penalty0 444--453, 2019.

\bibitem[Dhakan et~al.(2019)Dhakan, Maji, Sharma, Saxena, Pulikkan, Grace, Gomez, Scaria, Amato, and Sharma]{dhakan2019unique}
Dhakan, D., Maji, A., Sharma, A., Saxena, R., Pulikkan, J., Grace, T., Gomez, A., Scaria, J., Amato, K., and Sharma, V.
\newblock The unique composition of indian gut microbiome, gene catalogue, and associated fecal metabolome deciphered using multi-omics approaches.
\newblock \emph{Gigascience}, 8\penalty0 (3):\penalty0 giz004, 2019.

\bibitem[Durack \& Lynch(2019)Durack and Lynch]{durack2019gut}
Durack, J. and Lynch, S.~V.
\newblock The gut microbiome: Relationships with disease and opportunities for therapy.
\newblock \emph{Journal of experimental medicine}, 216\penalty0 (1):\penalty0 20--40, 2019.

\bibitem[Feng et~al.(2015)Feng, Liang, Jia, Stadlmayr, Tang, Lan, Zhang, Xia, Xu, Jie, et~al.]{feng2015gut}
Feng, Q., Liang, S., Jia, H., Stadlmayr, A., Tang, L., Lan, Z., Zhang, D., Xia, H., Xu, X., Jie, Z., et~al.
\newblock Gut microbiome development along the colorectal adenoma--carcinoma sequence.
\newblock \emph{Nature communications}, 6\penalty0 (1):\penalty0 6528, 2015.

\bibitem[Ferretti et~al.(2018)Ferretti, Pasolli, Tett, Asnicar, Gorfer, Fedi, Armanini, Truong, Manara, Zolfo, et~al.]{ferretti2018mother}
Ferretti, P., Pasolli, E., Tett, A., Asnicar, F., Gorfer, V., Fedi, S., Armanini, F., Truong, D.~T., Manara, S., Zolfo, M., et~al.
\newblock Mother-to-infant microbial transmission from different body sites shapes the developing infant gut microbiome.
\newblock \emph{Cell host \& microbe}, 24\penalty0 (1):\penalty0 133--145, 2018.

\bibitem[Fong \& Motani(2024)Fong and Motani]{sredt_Fong_Motani_2024}
Fong, K.~S. and Motani, M.
\newblock Symbolic regression enhanced decision trees for classification tasks.
\newblock \emph{Proceedings of the AAAI Conference on Artificial Intelligence}, 38\penalty0 (11):\penalty0 12033--12042, Mar. 2024.
\newblock \doi{10.1609/aaai.v38i11.29091}.
\newblock URL \url{https://ojs.aaai.org/index.php/AAAI/article/view/29091}.

\bibitem[Gomaa(2020)]{gomaa2020human}
Gomaa, E.~Z.
\newblock Human gut microbiota/microbiome in health and diseases: a review.
\newblock \emph{Antonie Van Leeuwenhoek}, 113\penalty0 (12):\penalty0 2019--2040, 2020.

\bibitem[Gordon-Rodriguez et~al.(2022)Gordon-Rodriguez, Quinn, and Cunningham]{gordon2022data}
Gordon-Rodriguez, E., Quinn, T., and Cunningham, J.~P.
\newblock Data augmentation for compositional data: Advancing predictive models of the microbiome.
\newblock \emph{Advances in Neural Information Processing Systems}, 35:\penalty0 20551--20565, 2022.

\bibitem[Gupta et~al.(2019)Gupta, Youn, Shin, and Suk]{gupta2019role}
Gupta, H., Youn, G.~S., Shin, M.~J., and Suk, K.~T.
\newblock Role of gut microbiota in hepatocarcinogenesis.
\newblock \emph{Microorganisms}, 7\penalty0 (5):\penalty0 121, 2019.

\bibitem[Hall et~al.(2017)Hall, Yassour, Sauk, Garner, Jiang, Arthur, Lagoudas, Vatanen, Fornelos, Wilson, et~al.]{hall2017novel}
Hall, A.~B., Yassour, M., Sauk, J., Garner, A., Jiang, X., Arthur, T., Lagoudas, G.~K., Vatanen, T., Fornelos, N., Wilson, R., et~al.
\newblock A novel ruminococcus gnavus clade enriched in inflammatory bowel disease patients.
\newblock \emph{Genome medicine}, 9:\penalty0 1--12, 2017.

\bibitem[Hannigan et~al.(2017)Hannigan, Duhaime, Ruffin~IV, Koumpouras, and Schloss]{hannigan2017viral}
Hannigan, G.~D., Duhaime, M.~B., Ruffin~IV, M.~T., Koumpouras, C.~C., and Schloss, P.~D.
\newblock Viral and bacterial communities of colorectal cancer.
\newblock \emph{BioRxiv}, pp.\  152868, 2017.

\bibitem[Hansen et~al.(2018)Hansen, Roager, S{\o}ndertoft, G{\o}bel, Kristensen, Vall{\`e}s-Colomer, Vieira-Silva, Ibr{\"u}gger, Lind, M{\ae}rkedahl, et~al.]{hansen2018low}
Hansen, L.~B., Roager, H.~M., S{\o}ndertoft, N.~B., G{\o}bel, R.~J., Kristensen, M., Vall{\`e}s-Colomer, M., Vieira-Silva, S., Ibr{\"u}gger, S., Lind, M.~V., M{\ae}rkedahl, R.~B., et~al.
\newblock A low-gluten diet induces changes in the intestinal microbiome of healthy danish adults.
\newblock \emph{Nature communications}, 9\penalty0 (1):\penalty0 4630, 2018.

\bibitem[Heintz-Buschart et~al.(2016)Heintz-Buschart, May, Laczny, Lebrun, Bellora, Krishna, Wampach, Schneider, Hogan, De~Beaufort, et~al.]{heintz2016integrated}
Heintz-Buschart, A., May, P., Laczny, C.~C., Lebrun, L.~A., Bellora, C., Krishna, A., Wampach, L., Schneider, J.~G., Hogan, A., De~Beaufort, C., et~al.
\newblock Integrated multi-omics of the human gut microbiome in a case study of familial type 1 diabetes.
\newblock \emph{Nature microbiology}, 2\penalty0 (1):\penalty0 1--13, 2016.

\bibitem[Hern{\'a}ndez~Medina et~al.(2022)Hern{\'a}ndez~Medina, Kutuzova, Nielsen, Johansen, Hansen, Nielsen, and Rasmussen]{hernandez2022machine}
Hern{\'a}ndez~Medina, R., Kutuzova, S., Nielsen, K.~N., Johansen, J., Hansen, L.~H., Nielsen, M., and Rasmussen, S.
\newblock Machine learning and deep learning applications in microbiome research.
\newblock \emph{ISME communications}, 2\penalty0 (1):\penalty0 98, 2022.

\bibitem[Ianiro et~al.(2022)Ianiro, Iorio, Porcari, Masucci, Sanguinetti, Perno, Gasbarrini, Putignani, and Cammarota]{ianiro2022gut}
Ianiro, G., Iorio, A., Porcari, S., Masucci, L., Sanguinetti, M., Perno, C.~F., Gasbarrini, A., Putignani, L., and Cammarota, G.
\newblock How the gut parasitome affects human health.
\newblock \emph{Therapeutic advances in gastroenterology}, 15:\penalty0 17562848221091524, 2022.

\bibitem[Ijaz et~al.(2017)Ijaz, Quince, Hanske, Loman, Calus, Bertz, Edwards, Gaya, Hansen, McGrogan, et~al.]{ijaz2017distinct}
Ijaz, U.~Z., Quince, C., Hanske, L., Loman, N., Calus, S.~T., Bertz, M., Edwards, C.~A., Gaya, D.~R., Hansen, R., McGrogan, P., et~al.
\newblock The distinct features of microbial ‘dysbiosis’ of crohn’s disease do not occur to the same extent in their unaffected, genetically-linked kindred.
\newblock \emph{PloS one}, 12\penalty0 (2):\penalty0 e0172605, 2017.

\bibitem[Jie et~al.(2017)Jie, Xia, Zhong, Feng, Li, Liang, Zhong, Liu, Gao, Zhao, et~al.]{jie2017gut}
Jie, Z., Xia, H., Zhong, S.-L., Feng, Q., Li, S., Liang, S., Zhong, H., Liu, Z., Gao, Y., Zhao, H., et~al.
\newblock The gut microbiome in atherosclerotic cardiovascular disease.
\newblock \emph{Nature communications}, 8\penalty0 (1):\penalty0 845, 2017.

\bibitem[Kamienny et~al.(2022)Kamienny, d'Ascoli, Lample, and Charton]{kamienny2022end}
Kamienny, P.-A., d'Ascoli, S., Lample, G., and Charton, F.
\newblock End-to-end symbolic regression with transformers.
\newblock \emph{Advances in Neural Information Processing Systems}, 35:\penalty0 10269--10281, 2022.

\bibitem[Karlsson et~al.(2013)Karlsson, Tremaroli, Nookaew, Bergstr{\"o}m, Behre, Fagerberg, Nielsen, and B{\"a}ckhed]{karlsson2013gut}
Karlsson, F.~H., Tremaroli, V., Nookaew, I., Bergstr{\"o}m, G., Behre, C.~J., Fagerberg, B., Nielsen, J., and B{\"a}ckhed, F.
\newblock Gut metagenome in european women with normal, impaired and diabetic glucose control.
\newblock \emph{Nature}, 498\penalty0 (7452):\penalty0 99--103, 2013.

\bibitem[Kaur et~al.(2020)Kaur, Khatri, Akhtar, Subramanian, and Ramya]{kaur2020metagenomics}
Kaur, K., Khatri, I., Akhtar, A., Subramanian, S., and Ramya, T.
\newblock Metagenomics analysis reveals features unique to indian distal gut microbiota.
\newblock \emph{PloS one}, 15\penalty0 (4):\penalty0 e0231197, 2020.

\bibitem[Keohane et~al.(2020)Keohane, Ghosh, Jeffery, Molloy, O’Toole, and Shanahan]{keohane2020microbiome}
Keohane, D.~M., Ghosh, T.~S., Jeffery, I.~B., Molloy, M.~G., O’Toole, P.~W., and Shanahan, F.
\newblock Microbiome and health implications for ethnic minorities after enforced lifestyle changes.
\newblock \emph{Nature Medicine}, 26\penalty0 (7):\penalty0 1089--1095, 2020.

\bibitem[Kieser et~al.(2018)Kieser, Sarker, Sakwinska, Foata, Sultana, Khan, Islam, Porta, Combremont, Betrisey, et~al.]{kieser2018bangladeshi}
Kieser, S., Sarker, S.~A., Sakwinska, O., Foata, F., Sultana, S., Khan, Z., Islam, S., Porta, N., Combremont, S., Betrisey, B., et~al.
\newblock Bangladeshi children with acute diarrhoea show faecal microbiomes with increased streptococcus abundance, irrespective of diarrhoea aetiology.
\newblock \emph{Environmental microbiology}, 20\penalty0 (6):\penalty0 2256--2269, 2018.

\bibitem[Korns(2018)]{korns2018gpsr_classification}
Korns, M.~F.
\newblock An evolutionary algorithm for big data multi-class classification problems.
\newblock \emph{Genetic Programming Theory and Practice XIV}, pp.\  165--178, 2018.

\bibitem[Kostic et~al.(2015)Kostic, Gevers, Siljander, Vatanen, Hy{\"o}tyl{\"a}inen, H{\"a}m{\"a}l{\"a}inen, Peet, Tillmann, P{\"o}h{\"o}, Mattila, et~al.]{kostic2015dynamics}
Kostic, A.~D., Gevers, D., Siljander, H., Vatanen, T., Hy{\"o}tyl{\"a}inen, T., H{\"a}m{\"a}l{\"a}inen, A.-M., Peet, A., Tillmann, V., P{\"o}h{\"o}, P., Mattila, I., et~al.
\newblock The dynamics of the human infant gut microbiome in development and in progression toward type 1 diabetes.
\newblock \emph{Cell host \& microbe}, 17\penalty0 (2):\penalty0 260--273, 2015.

\bibitem[Koza(1994)]{koza1994geneticprogramming}
Koza, J.~R.
\newblock Genetic programming as a means for programming computers by natural selection.
\newblock \emph{Statistics and computing}, 4:\penalty0 87--112, 1994.

\bibitem[Kurilshikov et~al.(2019)Kurilshikov, van~den Munckhof, Chen, Bonder, Schraa, Rutten, Riksen, de~Graaf, Oosting, Sanna, et~al.]{kurilshikov2019gut}
Kurilshikov, A., van~den Munckhof, I.~C., Chen, L., Bonder, M.~J., Schraa, K., Rutten, J.~H., Riksen, N.~P., de~Graaf, J., Oosting, M., Sanna, S., et~al.
\newblock Gut microbial associations to plasma metabolites linked to cardiovascular phenotypes and risk: a cross-sectional study.
\newblock \emph{Circulation research}, 124\penalty0 (12):\penalty0 1808--1820, 2019.

\bibitem[Le~Chatelier et~al.(2013)Le~Chatelier, Nielsen, Qin, Prifti, Hildebrand, Falony, Almeida, Arumugam, Batto, Kennedy, et~al.]{le2013richness}
Le~Chatelier, E., Nielsen, T., Qin, J., Prifti, E., Hildebrand, F., Falony, G., Almeida, M., Arumugam, M., Batto, J.-M., Kennedy, S., et~al.
\newblock Richness of human gut microbiome correlates with metabolic markers.
\newblock \emph{Nature}, 500\penalty0 (7464):\penalty0 541--546, 2013.

\bibitem[Li et~al.(2014)Li, Jia, Cai, Zhong, Feng, Sunagawa, Arumugam, Kultima, Prifti, Nielsen, et~al.]{li2014integrated}
Li, J., Jia, H., Cai, X., Zhong, H., Feng, Q., Sunagawa, S., Arumugam, M., Kultima, J.~R., Prifti, E., Nielsen, T., et~al.
\newblock An integrated catalog of reference genes in the human gut microbiome.
\newblock \emph{Nature biotechnology}, 32\penalty0 (8):\penalty0 834--841, 2014.

\bibitem[Li et~al.(2017)Li, Zhao, Wang, Chen, Tao, Tian, Wu, Liu, Cui, Geng, et~al.]{li2017gut}
Li, J., Zhao, F., Wang, Y., Chen, J., Tao, J., Tian, G., Wu, S., Liu, W., Cui, Q., Geng, B., et~al.
\newblock Gut microbiota dysbiosis contributes to the development of hypertension.
\newblock \emph{Microbiome}, 5:\penalty0 1--19, 2017.

\bibitem[Li et~al.(2016)Li, Zhu, Benes, Costea, Hercog, Hildebrand, Huerta-Cepas, Nieuwdorp, Saloj{\"a}rvi, Voigt, et~al.]{li2016durable}
Li, S.~S., Zhu, A., Benes, V., Costea, P.~I., Hercog, R., Hildebrand, F., Huerta-Cepas, J., Nieuwdorp, M., Saloj{\"a}rvi, J., Voigt, A.~Y., et~al.
\newblock Durable coexistence of donor and recipient strains after fecal microbiota transplantation.
\newblock \emph{Science}, 352\penalty0 (6285):\penalty0 586--589, 2016.

\bibitem[Liu et~al.(2016)Liu, Zhang, Wu, Cai, Huang, Chen, Xi, Liang, Hou, Zhou, et~al.]{liu2016unique}
Liu, W., Zhang, J., Wu, C., Cai, S., Huang, W., Chen, J., Xi, X., Liang, Z., Hou, Q., Zhou, B., et~al.
\newblock Unique features of ethnic mongolian gut microbiome revealed by metagenomic analysis.
\newblock \emph{Scientific reports}, 6\penalty0 (1):\penalty0 34826, 2016.

\bibitem[Lloyd-Price et~al.(2019)Lloyd-Price, Arze, Ananthakrishnan, Schirmer, Avila-Pacheco, Poon, Andrews, Ajami, Bonham, Brislawn, et~al.]{lloyd2019multi}
Lloyd-Price, J., Arze, C., Ananthakrishnan, A.~N., Schirmer, M., Avila-Pacheco, J., Poon, T.~W., Andrews, E., Ajami, N.~J., Bonham, K.~S., Brislawn, C.~J., et~al.
\newblock Multi-omics of the gut microbial ecosystem in inflammatory bowel diseases.
\newblock \emph{Nature}, 569\penalty0 (7758):\penalty0 655--662, 2019.

\bibitem[Lokmer et~al.(2019)Lokmer, Cian, Froment, Gantois, Viscogliosi, Chab{\'e}, and Segurel]{lokmer2019use}
Lokmer, A., Cian, A., Froment, A., Gantois, N., Viscogliosi, E., Chab{\'e}, M., and Segurel, L.
\newblock Use of shotgun metagenomics for the identification of protozoa in the gut microbiota of healthy individuals from worldwide populations with various industrialization levels.
\newblock \emph{PloS one}, 14\penalty0 (2):\penalty0 e0211139, 2019.

\bibitem[Louis et~al.(2016)Louis, Tappu, Damms-Machado, Huson, and Bischoff]{louis2016characterization}
Louis, S., Tappu, R.-M., Damms-Machado, A., Huson, D.~H., and Bischoff, S.~C.
\newblock Characterization of the gut microbial community of obese patients following a weight-loss intervention using whole metagenome shotgun sequencing.
\newblock \emph{PLoS One}, 11\penalty0 (2):\penalty0 e0149564, 2016.

\bibitem[L{\"o}wenmark et~al.(2020)L{\"o}wenmark, L{\"o}fgren-Burstr{\"o}m, Zingmark, Ekl{\"o}f, Dahlberg, Wai, Larsson, Ljuslinder, Edin, and Palmqvist]{lowenmark2020parvimonas}
L{\"o}wenmark, T., L{\"o}fgren-Burstr{\"o}m, A., Zingmark, C., Ekl{\"o}f, V., Dahlberg, M., Wai, S.~N., Larsson, P., Ljuslinder, I., Edin, S., and Palmqvist, R.
\newblock Parvimonas micra as a putative non-invasive faecal biomarker for colorectal cancer.
\newblock \emph{Scientific reports}, 10\penalty0 (1):\penalty0 15250, 2020.

\bibitem[Lundberg \& Lee(2017)Lundberg and Lee]{lundberg2017unified}
Lundberg, S.~M. and Lee, S.-I.
\newblock A unified approach to interpreting model predictions.
\newblock \emph{Advances in neural information processing systems}, 30, 2017.

\bibitem[Mehta et~al.(2018)Mehta, Abu-Ali, Drew, Lloyd-Price, Subramanian, Lochhead, Joshi, Ivey, Khalili, Brown, et~al.]{mehta2018stability}
Mehta, R.~S., Abu-Ali, G.~S., Drew, D.~A., Lloyd-Price, J., Subramanian, A., Lochhead, P., Joshi, A.~D., Ivey, K.~L., Khalili, H., Brown, G.~T., et~al.
\newblock Stability of the human faecal microbiome in a cohort of adult men.
\newblock \emph{Nature microbiology}, 3\penalty0 (3):\penalty0 347--355, 2018.

\bibitem[Nagy-Szakal et~al.(2017)Nagy-Szakal, Williams, Mishra, Che, Lee, Bateman, Klimas, Komaroff, Levine, Montoya, et~al.]{nagy2017fecal}
Nagy-Szakal, D., Williams, B.~L., Mishra, N., Che, X., Lee, B., Bateman, L., Klimas, N.~G., Komaroff, A.~L., Levine, S., Montoya, J.~G., et~al.
\newblock Fecal metagenomic profiles in subgroups of patients with myalgic encephalomyelitis/chronic fatigue syndrome.
\newblock \emph{Microbiome}, 5:\penalty0 1--17, 2017.

\bibitem[Neu et~al.(2021)Neu, Allen, and Roy]{neu2021defining}
Neu, A.~T., Allen, E.~E., and Roy, K.
\newblock Defining and quantifying the core microbiome: challenges and prospects.
\newblock \emph{Proceedings of the National Academy of Sciences}, 118\penalty0 (51):\penalty0 e2104429118, 2021.

\bibitem[Nielsen(2014)]{nielsen2014systematic}
Nielsen, H.~B.
\newblock Systematic review of near-infrared spectroscopy determined cerebral oxygenation during non-cardiac surgery.
\newblock \emph{Frontiers in physiology}, 5:\penalty0 70501, 2014.

\bibitem[Obregon-Tito et~al.(2015)Obregon-Tito, Tito, Metcalf, Sankaranarayanan, Clemente, Ursell, Zech~Xu, Van~Treuren, Knight, Gaffney, et~al.]{obregon2015subsistence}
Obregon-Tito, A.~J., Tito, R.~Y., Metcalf, J., Sankaranarayanan, K., Clemente, J.~C., Ursell, L.~K., Zech~Xu, Z., Van~Treuren, W., Knight, R., Gaffney, P.~M., et~al.
\newblock Subsistence strategies in traditional societies distinguish gut microbiomes.
\newblock \emph{Nature communications}, 6\penalty0 (1):\penalty0 6505, 2015.

\bibitem[Osman et~al.(2021)Osman, Neoh, Ab~Mutalib, Chin, Mazlan, Raja~Ali, Zakaria, Ngiu, Ang, and Jamal]{osman2021parvimonas}
Osman, M.~A., Neoh, H.-m., Ab~Mutalib, N.-S., Chin, S.-F., Mazlan, L., Raja~Ali, R.~A., Zakaria, A.~D., Ngiu, C.~S., Ang, M.~Y., and Jamal, R.
\newblock Parvimonas micra, peptostreptococcus stomatis, fusobacterium nucleatum and akkermansia muciniphila as a four-bacteria biomarker panel of colorectal cancer.
\newblock \emph{Scientific reports}, 11\penalty0 (1):\penalty0 2925, 2021.

\bibitem[Pasolli et~al.(2017)Pasolli, Schiffer, Manghi, Renson, Obenchain, Truong, Beghini, Malik, Ramos, Dowd, Huttenhower, Morgan, Segata, and Waldron]{datasets}
Pasolli, E., Schiffer, L., Manghi, P., Renson, A., Obenchain, V., Truong, D.~T., Beghini, F., Malik, F., Ramos, M., Dowd, J.~B., Huttenhower, C., Morgan, M., Segata, N., and Waldron, L.
\newblock Accessible, curated metagenomic data through {ExperimentHub}.
\newblock \emph{Nat. Methods}, 14\penalty0 (11):\penalty0 1023--1024, oct 2017.
\newblock ISSN 1548-7091, 1548-7105.
\newblock \doi{10.1038/nmeth.4468}.

\bibitem[Pasolli et~al.(2019)Pasolli, Asnicar, Manara, Zolfo, Karcher, Armanini, Beghini, Manghi, Tett, Ghensi, et~al.]{pasolli2019extensive}
Pasolli, E., Asnicar, F., Manara, S., Zolfo, M., Karcher, N., Armanini, F., Beghini, F., Manghi, P., Tett, A., Ghensi, P., et~al.
\newblock Extensive unexplored human microbiome diversity revealed by over 150,000 genomes from metagenomes spanning age, geography, and lifestyle.
\newblock \emph{Cell}, 176\penalty0 (3):\penalty0 649--662, 2019.

\bibitem[Pedregosa et~al.(2011)Pedregosa, Varoquaux, Gramfort, Michel, Thirion, Grisel, Blondel, Prettenhofer, Weiss, Dubourg, et~al.]{scikit-learn}
Pedregosa, F., Varoquaux, G., Gramfort, A., Michel, V., Thirion, B., Grisel, O., Blondel, M., Prettenhofer, P., Weiss, R., Dubourg, V., et~al.
\newblock Scikit-learn: Machine learning in python.
\newblock \emph{the Journal of machine Learning research}, 12:\penalty0 2825--2830, 2011.

\bibitem[Pehrsson et~al.(2016)Pehrsson, Tsukayama, Patel, Mej{\'\i}a-Bautista, Sosa-Soto, Navarrete, Calderon, Cabrera, Hoyos-Arango, Bertoli, et~al.]{pehrsson2016interconnected}
Pehrsson, E.~C., Tsukayama, P., Patel, S., Mej{\'\i}a-Bautista, M., Sosa-Soto, G., Navarrete, K.~M., Calderon, M., Cabrera, L., Hoyos-Arango, W., Bertoli, M.~T., et~al.
\newblock Interconnected microbiomes and resistomes in low-income human habitats.
\newblock \emph{Nature}, 533\penalty0 (7602):\penalty0 212--216, 2016.

\bibitem[Petersen et~al.(2021)Petersen, Landajuela, Mundhenk, Santiago, Kim, and Kim]{petersen2021deepSR}
Petersen, B.~K., Landajuela, M., Mundhenk, T.~N., Santiago, C.~P., Kim, S.~K., and Kim, J.~T.
\newblock Deep symbolic regression: Recovering mathematical expressions from data via risk-seeking policy gradients, 2021.

\bibitem[Qin et~al.(2012)Qin, Li, Cai, Li, Zhu, Zhang, Liang, Zhang, Guan, Shen, et~al.]{qin2012metagenome}
Qin, J., Li, Y., Cai, Z., Li, S., Zhu, J., Zhang, F., Liang, S., Zhang, W., Guan, Y., Shen, D., et~al.
\newblock A metagenome-wide association study of gut microbiota in type 2 diabetes.
\newblock \emph{nature}, 490\penalty0 (7418):\penalty0 55--60, 2012.

\bibitem[Qin et~al.(2014)Qin, Yang, Li, Prifti, Chen, Shao, Guo, Le~Chatelier, Yao, Wu, et~al.]{qin2014alterations}
Qin, N., Yang, F., Li, A., Prifti, E., Chen, Y., Shao, L., Guo, J., Le~Chatelier, E., Yao, J., Wu, L., et~al.
\newblock Alterations of the human gut microbiome in liver cirrhosis.
\newblock \emph{Nature}, 513\penalty0 (7516):\penalty0 59--64, 2014.

\bibitem[Rampelli et~al.(2015)Rampelli, Schnorr, Consolandi, Turroni, Severgnini, Peano, Brigidi, Crittenden, Henry, and Candela]{rampelli2015metagenome}
Rampelli, S., Schnorr, S.~L., Consolandi, C., Turroni, S., Severgnini, M., Peano, C., Brigidi, P., Crittenden, A.~N., Henry, A.~G., and Candela, M.
\newblock Metagenome sequencing of the hadza hunter-gatherer gut microbiota.
\newblock \emph{Current Biology}, 25\penalty0 (13):\penalty0 1682--1693, 2015.

\bibitem[Raymond et~al.(2016)Raymond, Ouameur, D{\'e}raspe, Iqbal, Gingras, Dridi, Leprohon, Plante, Giroux, B{\'e}rub{\'e}, et~al.]{raymond2016initial}
Raymond, F., Ouameur, A.~A., D{\'e}raspe, M., Iqbal, N., Gingras, H., Dridi, B., Leprohon, P., Plante, P.-L., Giroux, R., B{\'e}rub{\'e}, {\`E}., et~al.
\newblock The initial state of the human gut microbiome determines its reshaping by antibiotics.
\newblock \emph{The ISME journal}, 10\penalty0 (3):\penalty0 707--720, 2016.

\bibitem[Ribeiro et~al.(2016)Ribeiro, Singh, and Guestrin]{ribeiro2016should}
Ribeiro, M.~T., Singh, S., and Guestrin, C.
\newblock " why should i trust you?" explaining the predictions of any classifier.
\newblock In \emph{Proceedings of the 22nd ACM SIGKDD international conference on knowledge discovery and data mining}, pp.\  1135--1144, 2016.

\bibitem[Rosa et~al.(2018)Rosa, Supali, Gankpala, Djuardi, Sartono, Zhou, Fischer, Martin, Tyagi, Bolay, et~al.]{rosa2018differential}
Rosa, B.~A., Supali, T., Gankpala, L., Djuardi, Y., Sartono, E., Zhou, Y., Fischer, K., Martin, J., Tyagi, R., Bolay, F.~K., et~al.
\newblock Differential human gut microbiome assemblages during soil-transmitted helminth infections in indonesia and liberia.
\newblock \emph{Microbiome}, 6:\penalty0 1--19, 2018.

\bibitem[Ruaud et~al.(2022)Ruaud, Pfister, Ley, and Youngblut]{ruaud2022endoR}
Ruaud, A., Pfister, N., Ley, R.~E., and Youngblut, N.~D.
\newblock Interpreting tree ensemble machine learning models with endor.
\newblock \emph{PLOS Computational Biology}, 18\penalty0 (12):\penalty0 e1010714, 2022.

\bibitem[Rubel et~al.(2020)Rubel, Abbas, Taylor, Connell, Tanes, Bittinger, Ndze, Fonsah, Ngwang, Essiane, et~al.]{rubel2020lifestyle}
Rubel, M.~A., Abbas, A., Taylor, L.~J., Connell, A., Tanes, C., Bittinger, K., Ndze, V.~N., Fonsah, J.~Y., Ngwang, E., Essiane, A., et~al.
\newblock Lifestyle and the presence of helminths is associated with gut microbiome composition in cameroonians.
\newblock \emph{Genome biology}, 21:\penalty0 1--32, 2020.

\bibitem[Sankaranarayanan et~al.(2015)Sankaranarayanan, Ozga, Warinner, Tito, Obregon-Tito, Xu, Gaffney, Jervis, Cox, Stephens, et~al.]{sankaranarayanan2015gut}
Sankaranarayanan, K., Ozga, A.~T., Warinner, C., Tito, R.~Y., Obregon-Tito, A.~J., Xu, J., Gaffney, P.~M., Jervis, L.~L., Cox, D., Stephens, L., et~al.
\newblock Gut microbiome diversity among cheyenne and arapaho individuals from western oklahoma.
\newblock \emph{Current Biology}, 25\penalty0 (24):\penalty0 3161--3169, 2015.

\bibitem[Sarwal et~al.(2022)Sarwal, Brito, Mangul, and Koslicki]{TAMPA_Sarwal2022}
Sarwal, V., Brito, J., Mangul, S., and Koslicki, D.
\newblock Tampa: interpretable analysis and visualization of metagenomics-based taxon abundance profiles.
\newblock \emph{bioRxiv}, 2022.
\newblock \doi{10.1101/2022.04.28.489926}.
\newblock URL \url{https://www.biorxiv.org/content/early/2022/04/29/2022.04.28.489926}.

\bibitem[Schirmer et~al.(2016)Schirmer, Smeekens, Vlamakis, Jaeger, Oosting, Franzosa, Ter~Horst, Jansen, Jacobs, Bonder, et~al.]{schirmer2016linking}
Schirmer, M., Smeekens, S.~P., Vlamakis, H., Jaeger, M., Oosting, M., Franzosa, E.~A., Ter~Horst, R., Jansen, T., Jacobs, L., Bonder, M.~J., et~al.
\newblock Linking the human gut microbiome to inflammatory cytokine production capacity.
\newblock \emph{Cell}, 167\penalty0 (4):\penalty0 1125--1136, 2016.

\bibitem[Schirmer et~al.(2018)Schirmer, Franzosa, Lloyd-Price, McIver, Schwager, Poon, Ananthakrishnan, Andrews, Barron, Lake, et~al.]{schirmer2018dynamics}
Schirmer, M., Franzosa, E.~A., Lloyd-Price, J., McIver, L.~J., Schwager, R., Poon, T.~W., Ananthakrishnan, A.~N., Andrews, E., Barron, G., Lake, K., et~al.
\newblock Dynamics of metatranscription in the inflammatory bowel disease gut microbiome.
\newblock \emph{Nature microbiology}, 3\penalty0 (3):\penalty0 337--346, 2018.

\bibitem[Schmidt \& Lipson(2009)Schmidt and Lipson]{schmidt2009SReureqa}
Schmidt, M. and Lipson, H.
\newblock Distilling free-form natural laws from experimental data.
\newblock \emph{science}, 324\penalty0 (5923):\penalty0 81--85, 2009.

\bibitem[Schmidt \& Lipson(2010)Schmidt and Lipson]{Eureqa_SRforEquations}
Schmidt, M. and Lipson, H.
\newblock \emph{Symbolic Regression of Implicit Equations}, pp.\  73--85.
\newblock Springer US, Boston, MA, 2010.
\newblock ISBN 978-1-4419-1626-6.
\newblock \doi{10.1007/978-1-4419-1626-6_5}.
\newblock URL \url{https://doi.org/10.1007/978-1-4419-1626-6_5}.

\bibitem[Shimomura et~al.(2023)Shimomura, Zha, Komukai, Narii, Sobue, Kitamura, Shiba, Mizutani, Yamada, Sawada, et~al.]{shimomura2023mediation}
Shimomura, Y., Zha, L., Komukai, S., Narii, N., Sobue, T., Kitamura, T., Shiba, S., Mizutani, S., Yamada, T., Sawada, N., et~al.
\newblock Mediation effect of intestinal microbiota on the relationship between fiber intake and colorectal cancer.
\newblock \emph{International Journal of Cancer}, 152\penalty0 (9):\penalty0 1752--1762, 2023.

\bibitem[Smits et~al.(2017)Smits, Leach, Sonnenburg, Gonzalez, Lichtman, Reid, Knight, Manjurano, Changalucha, Elias, et~al.]{smits2017seasonal}
Smits, S.~A., Leach, J., Sonnenburg, E.~D., Gonzalez, C.~G., Lichtman, J.~S., Reid, G., Knight, R., Manjurano, A., Changalucha, J., Elias, J.~E., et~al.
\newblock Seasonal cycling in the gut microbiome of the hadza hunter-gatherers of tanzania.
\newblock \emph{Science}, 357\penalty0 (6353):\penalty0 802--806, 2017.

\bibitem[Stephens()]{Stephens}
Stephens, T.
\newblock Welcome to gplearn’s documentation!¶.
\newblock URL \url{https://gplearn.readthedocs.io/en/stable/}.

\bibitem[Tett et~al.(2019)Tett, Huang, Asnicar, Fehlner-Peach, Pasolli, Karcher, Armanini, Manghi, Bonham, Zolfo, et~al.]{tett2019prevotella}
Tett, A., Huang, K.~D., Asnicar, F., Fehlner-Peach, H., Pasolli, E., Karcher, N., Armanini, F., Manghi, P., Bonham, K., Zolfo, M., et~al.
\newblock The prevotella copri complex comprises four distinct clades underrepresented in westernized populations.
\newblock \emph{Cell host \& microbe}, 26\penalty0 (5):\penalty0 666--679, 2019.

\bibitem[Thomas et~al.(2019)Thomas, Manghi, Asnicar, Pasolli, Armanini, Zolfo, Beghini, Manara, Karcher, Pozzi, et~al.]{thomas2019metagenomic}
Thomas, A.~M., Manghi, P., Asnicar, F., Pasolli, E., Armanini, F., Zolfo, M., Beghini, F., Manara, S., Karcher, N., Pozzi, C., et~al.
\newblock Metagenomic analysis of colorectal cancer datasets identifies cross-cohort microbial diagnostic signatures and a link with choline degradation.
\newblock \emph{Nature medicine}, 25\penalty0 (4):\penalty0 667--678, 2019.

\bibitem[Vatanen et~al.(2016)Vatanen, Kostic, d’Hennezel, Siljander, Franzosa, Yassour, Kolde, Vlamakis, Arthur, H{\"a}m{\"a}l{\"a}inen, et~al.]{vatanen2016variation}
Vatanen, T., Kostic, A.~D., d’Hennezel, E., Siljander, H., Franzosa, E.~A., Yassour, M., Kolde, R., Vlamakis, H., Arthur, T.~D., H{\"a}m{\"a}l{\"a}inen, A.-M., et~al.
\newblock Variation in microbiome lps immunogenicity contributes to autoimmunity in humans.
\newblock \emph{Cell}, 165\penalty0 (4):\penalty0 842--853, 2016.

\bibitem[Vieira-Silva et~al.(2020)Vieira-Silva, Falony, Belda, Nielsen, Aron-Wisnewsky, Chakaroun, Forslund, Assmann, Valles-Colomer, Nguyen, et~al.]{vieira2020statin}
Vieira-Silva, S., Falony, G., Belda, E., Nielsen, T., Aron-Wisnewsky, J., Chakaroun, R., Forslund, S.~K., Assmann, K., Valles-Colomer, M., Nguyen, T. T.~D., et~al.
\newblock Statin therapy is associated with lower prevalence of gut microbiota dysbiosis.
\newblock \emph{Nature}, 581\penalty0 (7808):\penalty0 310--315, 2020.

\bibitem[Vincent et~al.(2016)Vincent, Miller, Edens, Mehrotra, Dewar, and Manges]{vincent2016bloom}
Vincent, C., Miller, M.~A., Edens, T.~J., Mehrotra, S., Dewar, K., and Manges, A.~R.
\newblock Bloom and bust: intestinal microbiota dynamics in response to hospital exposures and clostridium difficile colonization or infection.
\newblock \emph{Microbiome}, 4:\penalty0 1--11, 2016.

\bibitem[Vogtmann et~al.(2016)Vogtmann, Hua, Zeller, Sunagawa, Voigt, Hercog, Goedert, Shi, Bork, and Sinha]{vogtmann2016colorectal}
Vogtmann, E., Hua, X., Zeller, G., Sunagawa, S., Voigt, A.~Y., Hercog, R., Goedert, J.~J., Shi, J., Bork, P., and Sinha, R.
\newblock Colorectal cancer and the human gut microbiome: reproducibility with whole-genome shotgun sequencing.
\newblock \emph{PloS one}, 11\penalty0 (5):\penalty0 e0155362, 2016.

\bibitem[Wampach et~al.(2018)Wampach, Heintz-Buschart, Fritz, Ramiro-Garcia, Habier, Herold, Narayanasamy, Kaysen, Hogan, Bindl, et~al.]{wampach2018birth}
Wampach, L., Heintz-Buschart, A., Fritz, J.~V., Ramiro-Garcia, J., Habier, J., Herold, M., Narayanasamy, S., Kaysen, A., Hogan, A.~H., Bindl, L., et~al.
\newblock Birth mode is associated with earliest strain-conferred gut microbiome functions and immunostimulatory potential.
\newblock \emph{Nature communications}, 9\penalty0 (1):\penalty0 5091, 2018.

\bibitem[Wirbel et~al.(2019)Wirbel, Pyl, Kartal, Zych, Kashani, Milanese, Fleck, Voigt, Palleja, Ponnudurai, et~al.]{wirbel2019meta}
Wirbel, J., Pyl, P.~T., Kartal, E., Zych, K., Kashani, A., Milanese, A., Fleck, J.~S., Voigt, A.~Y., Palleja, A., Ponnudurai, R., et~al.
\newblock Meta-analysis of fecal metagenomes reveals global microbial signatures that are specific for colorectal cancer.
\newblock \emph{Nature medicine}, 25\penalty0 (4):\penalty0 679--689, 2019.

\bibitem[Xie et~al.(2016)Xie, Guo, Zhong, Feng, Lan, Qin, Ward, Jackson, Xia, Chen, et~al.]{xie2016shotgun}
Xie, H., Guo, R., Zhong, H., Feng, Q., Lan, Z., Qin, B., Ward, K.~J., Jackson, M.~A., Xia, Y., Chen, X., et~al.
\newblock Shotgun metagenomics of 250 adult twins reveals genetic and environmental impacts on the gut microbiome.
\newblock \emph{Cell systems}, 3\penalty0 (6):\penalty0 572--584, 2016.

\bibitem[Yachida et~al.(2019)Yachida, Mizutani, Shiroma, Shiba, Nakajima, Sakamoto, Watanabe, Masuda, Nishimoto, Kubo, et~al.]{yachida2019metagenomic}
Yachida, S., Mizutani, S., Shiroma, H., Shiba, S., Nakajima, T., Sakamoto, T., Watanabe, H., Masuda, K., Nishimoto, Y., Kubo, M., et~al.
\newblock Metagenomic and metabolomic analyses reveal distinct stage-specific phenotypes of the gut microbiota in colorectal cancer.
\newblock \emph{Nature medicine}, 25\penalty0 (6):\penalty0 968--976, 2019.

\bibitem[Yassour et~al.(2016)Yassour, Lim, Yun, Tickle, Sung, Song, Lee, Franzosa, Morgan, Gevers, et~al.]{yassour2016sub}
Yassour, M., Lim, M.~Y., Yun, H.~S., Tickle, T.~L., Sung, J., Song, Y.-M., Lee, K., Franzosa, E.~A., Morgan, X.~C., Gevers, D., et~al.
\newblock Sub-clinical detection of gut microbial biomarkers of obesity and type 2 diabetes.
\newblock \emph{Genome medicine}, 8:\penalty0 1--14, 2016.

\bibitem[Yassour et~al.(2018)Yassour, Jason, Hogstrom, Arthur, Tripathi, Siljander, Selvenius, Oikarinen, Hy{\"o}ty, Virtanen, et~al.]{yassour2018strain}
Yassour, M., Jason, E., Hogstrom, L.~J., Arthur, T.~D., Tripathi, S., Siljander, H., Selvenius, J., Oikarinen, S., Hy{\"o}ty, H., Virtanen, S.~M., et~al.
\newblock Strain-level analysis of mother-to-child bacterial transmission during the first few months of life.
\newblock \emph{Cell host \& microbe}, 24\penalty0 (1):\penalty0 146--154, 2018.

\bibitem[Ye et~al.(2018)Ye, Zhang, Wu, Zhang, Wang, Huang, Du, Cao, Tang, Zhou, et~al.]{ye2018metagenomic}
Ye, Z., Zhang, N., Wu, C., Zhang, X., Wang, Q., Huang, X., Du, L., Cao, Q., Tang, J., Zhou, C., et~al.
\newblock A metagenomic study of the gut microbiome in behcet’s disease.
\newblock \emph{Microbiome}, 6:\penalty0 1--13, 2018.

\bibitem[Yu et~al.(2017)Yu, Feng, Wong, Zhang, yi~Liang, Qin, Tang, Zhao, Stenvang, Li, et~al.]{yu2017metagenomic}
Yu, J., Feng, Q., Wong, S.~H., Zhang, D., yi~Liang, Q., Qin, Y., Tang, L., Zhao, H., Stenvang, J., Li, Y., et~al.
\newblock Metagenomic analysis of faecal microbiome as a tool towards targeted non-invasive biomarkers for colorectal cancer.
\newblock \emph{Gut}, 66\penalty0 (1):\penalty0 70--78, 2017.

\bibitem[Zeevi et~al.(2015)Zeevi, Korem, Zmora, Israeli, Rothschild, Weinberger, Ben-Yacov, Lador, Avnit-Sagi, Lotan-Pompan, et~al.]{zeevi2015personalized}
Zeevi, D., Korem, T., Zmora, N., Israeli, D., Rothschild, D., Weinberger, A., Ben-Yacov, O., Lador, D., Avnit-Sagi, T., Lotan-Pompan, M., et~al.
\newblock Personalized nutrition by prediction of glycemic responses.
\newblock \emph{Cell}, 163\penalty0 (5):\penalty0 1079--1094, 2015.

\bibitem[Zeller et~al.(2014)Zeller, Tap, Voigt, Sunagawa, Kultima, Costea, Amiot, B{\"o}hm, Brunetti, Habermann, et~al.]{zeller2014potential}
Zeller, G., Tap, J., Voigt, A.~Y., Sunagawa, S., Kultima, J.~R., Costea, P.~I., Amiot, A., B{\"o}hm, J., Brunetti, F., Habermann, N., et~al.
\newblock Potential of fecal microbiota for early-stage detection of colorectal cancer.
\newblock \emph{Molecular systems biology}, 10\penalty0 (11):\penalty0 766, 2014.

\bibitem[Zhao et~al.(2022)Zhao, Zhang, Zhou, Fu, Lau, Chun, Cheung, Coker, Wei, Wu, et~al.]{zhao2022parvimonas}
Zhao, L., Zhang, X., Zhou, Y., Fu, K., Lau, H. C.-H., Chun, T. W.-Y., Cheung, A. H.-K., Coker, O.~O., Wei, H., Wu, W. K.-K., et~al.
\newblock Parvimonas micra promotes colorectal tumorigenesis and is associated with prognosis of colorectal cancer patients.
\newblock \emph{Oncogene}, 41\penalty0 (36):\penalty0 4200--4210, 2022.

\bibitem[Zhernakova et~al.(2016)Zhernakova, Kurilshikov, Bonder, Tigchelaar, Schirmer, Vatanen, Mujagic, Vila, Falony, Vieira-Silva, et~al.]{zhernakova2016population}
Zhernakova, A., Kurilshikov, A., Bonder, M.~J., Tigchelaar, E.~F., Schirmer, M., Vatanen, T., Mujagic, Z., Vila, A.~V., Falony, G., Vieira-Silva, S., et~al.
\newblock Population-based metagenomics analysis reveals markers for gut microbiome composition and diversity.
\newblock \emph{Science}, 352\penalty0 (6285):\penalty0 565--569, 2016.

\bibitem[Zhou et~al.(2019)Zhou, Sailani, Contrepois, Zhou, Ahadi, Leopold, Zhang, Rao, Avina, Mishra, et~al.]{zhou2019longitudinal}
Zhou, W., Sailani, M.~R., Contrepois, K., Zhou, Y., Ahadi, S., Leopold, S.~R., Zhang, M.~J., Rao, V., Avina, M., Mishra, T., et~al.
\newblock Longitudinal multi-omics of host--microbe dynamics in prediabetes.
\newblock \emph{Nature}, 569\penalty0 (7758):\penalty0 663--671, 2019.

\bibitem[Zhu et~al.(2020)Zhu, Ju, Wang, Wang, Guo, Ma, Sun, Fan, Xie, Yang, et~al.]{zhu2020metagenome}
Zhu, F., Ju, Y., Wang, W., Wang, Q., Guo, R., Ma, Q., Sun, Q., Fan, Y., Xie, Y., Yang, Z., et~al.
\newblock Metagenome-wide association of gut microbiome features for schizophrenia.
\newblock \emph{Nature communications}, 11\penalty0 (1):\penalty0 1612, 2020.

\end{thebibliography}
\bibliographystyle{icml2024}

\newpage
\appendix
\onecolumn

\section{Datasets}
\label{ap_datasets}

% https://www.biorxiv.org/content/10.1101/2023.10.25.563937v1.full.pdf
%https://github.com/waldronlab/curatedMetagenomicData/blob/3e51dcdaedb1a024920ff6f9a61c33888cc61398/NEWS.md?plain=1#L101
% number of samples per study 

Table \ref{tab:studies} provides a general overview of the all $71$ studies incorporated into our combined dataset. We specifically selected samples classified as either healthy or colorectal cancer (CRC). The resulting dataset comprises $11,137$ samples, with only $664$ samples belonging to the positive CRC class. This class imbalance adds significant complexity to the problem, underscoring the challenge of accurately predicting CRC presence. In our experiments, all models except XGBoost performed poorly ($F_1$ score in range $0.29$-$0.35$) on the entire imbalanced dataset, hence we decided to randomly undersample the healthy class.
All baseline models received the same normalized data versions.

\section{Reproducibility}
We utilized open-source Python libraries for the presented analysis. 
% In the case of acceptance, we will release our code and provide additional details to ensure full reproducibility of our results.
Our code and additional details ro reproduce the results described in the paper can be found in this github repository: \texttt{https://github.com/swag2198/microbiome-symbolic-regression}.

\section{Details of the models used and hyperparameters}
\label{appdx:models}
Table~\ref{tab:model-params} provide details about the ML models used in our experiments, and also mention the hyperparameters used in fitting the Symbolic Classifiers from \texttt{gplearn} library.

\begin{table}[h]
\centering

\caption{Models used and their code instantiations.}

\begin{tabular}{ll}
\toprule
\textbf{Models} &    \textbf{Parameters} \\
\midrule
   LR & \texttt{LogisticRegression(max\_iter=500)} \\
   DT & \texttt{DecisionTreeClassifier(max\_depth=5)} \\
   RF & \texttt{RandomForestClassifier(n\_estimators=50)} \\
   XG & \begin{tabular}{@{}c@{}}\texttt{XGBClassifier(n\_estimators=50, max\_depth=5,} \\ \texttt{learning\_rate=0.1, objective='binary:logistic')}\end{tabular}  \\
   SR & \begin{tabular}{@{}c@{}}\texttt{SymbolicClassifier(population\_size=6000,generations=20,tournament\_size=25,} \\ \texttt{init\_depth=(2, 6), parsimony\_coefficient=0.001)}\end{tabular}  \\   
\bottomrule
\end{tabular}

\label{tab:model-params}
\end{table}

\section{Example of SR and SRf expression}
\label{appdx:sr-vs-srf}

\begin{figure}
%\vspace{-1cm}
    \centering
    \includegraphics[width=0.4\columnwidth]{figures/a.pdf}   \includegraphics[width=0.4\columnwidth]{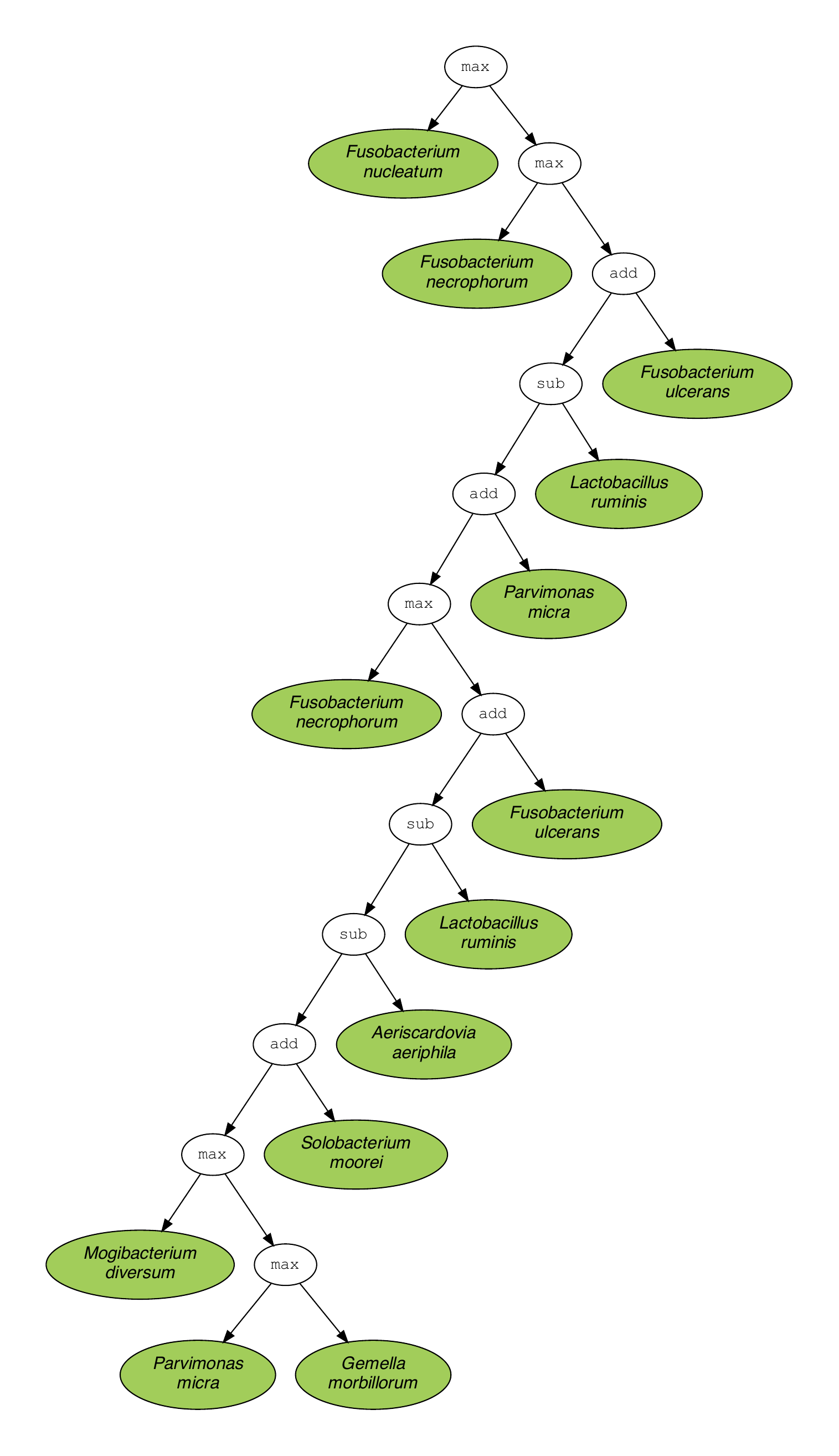}
    \caption{\small{
    (\textbf{Left}) We see an SRf expression learned using \texttt{gplearn}. The added custom function \texttt{ifelse} helps to reduce its size down to 15 nodes while the SR expression for the same run had 25 nodes (\textbf{Right}).
    }
    }
    \label{fig:sr-vs-srf}
    %\vspace{-.5cm}
\end{figure}

Fig. \ref{fig:sr-vs-srf} provides additional examples of obtained SR models. 

\begin{table}
\centering
\caption{Overview of studies included in the analysis.}
\scriptsize 
\begin{tabular}{lcc}
\toprule
\textbf{\texttt{CuratedMetagenomicData} ID} & \textbf{Number of Samples} & \textbf{Source}  \\
\midrule
KarlssonFH\_2013 & $145$ & \citet{karlsson2013gut} \\ 
RubelMA\_2020 & $175$  & \citet{rubel2020lifestyle} \\ 
AsnicarF\_2017 & $24$ & \citet{asnicar2017studying} \\ 
AsnicarF\_2021 & $1098$ & \citet{asnicar2021microbiome} \\
KaurK\_2020 & $31$ & \citet{kaur2020metagenomics} \\ 
SankaranarayananK\_2015 & $37$ & \citet{sankaranarayanan2015gut} \\ 
KeohaneDM\_2020 & $117$ & \citet{keohane2020microbiome} \\ 
SchirmerM\_2016 & $471$ & \citet{schirmer2016linking} \\ 
KieserS\_2018 & $27$ & \citet{kieser2018bangladeshi} \\
BedarfJR\_2017 & $59$ & \citet{bedarf2017functional} \\
KosticAD\_2015 & $120$ & \citet{kostic2015dynamics} \\ 
SmitsSA\_2017 & $40$ & \citet{smits2017seasonal} \\ 
Bengtsson-PalmeJ\_2015 & $70$ & \citet{bengtsson2015metaxa2} \\ 
LeChatelierE\_2013 & $292$ & \citet{le2013richness} \\
TettAJ\_2019\_a &  $68$ & \citet{tett2019prevotella} \\ 
BritoIL\_2016 & $312$ & \citet{brito2016mobile} \\ 
LifeLinesDeep\_2016 & $1135$ & \citet{kurilshikov2019gut} \\ 
BrooksB\_2017 & $408$ & \citet{zhernakova2016population} \\ 
LiJ\_2014 & $260$  & \citet{li2014integrated} \\ 
LiJ\_2017 & $196$ &\citet{li2017gut} \\ 
ChuDM\_2017 &  $86$ &\citet{chu2017maturation} \\ 
ThomasAM\_2018a & $80$ &\citet{thomas2019metagenomic} \\ 
ThomasAM\_2018b &  $60$ &\citet{thomas2019metagenomic} \\ 
ThomasAM\_2019\_c & $80$ &\citet{thomas2019metagenomic} \\ 
CosteaPI\_2017 & $279$ &\citet{costea2017towards} \\ 
LiSS\_2016 & $55$  &\citet{li2016durable} \\ 
DavidLA\_2015 & $47$ &\citet{david2015gut} \\ 
LiuW\_2016 & $110$ &\citet{liu2016unique} \\
DeFilippisF\_2019 & $97$ &\citet{de2019distinct} \\ 
LokmerA\_2019 & $57$ &\citet{lokmer2019use} \\ 
VatanenT\_2016 & $785$  &\citet{vatanen2016variation} \\ 
DhakanDB\_2019 & $110$ &\citet{dhakan2019unique} \\ 
LouisS\_2016 & $92$ &\citet{louis2016characterization} \\ 
VincentC\_2016 & $229$  &\citet{vincent2016bloom} \\ 
FengQ\_2015 & $154$  &\citet{feng2015gut} \\
MehtaRS\_2018 & $928$ &\citet{mehta2018stability} \\ 
VogtmannE\_2016 & $110$  &\citet{vogtmann2016colorectal} \\ 
FerrettiP\_2018 & $214$ &\citet{ferretti2018mother} \\ 
MetaCardis\_2020\_a & $1831$ &\citet{vieira2020statin} \\
WampachL\_2018 & $63$ &\citet{wampach2018birth} \\ 
GuptaA\_2019 & $60$ & \citet{gupta2019role} \\ 
NagySzakalD\_2017 & $396$ &\citet{nagy2017fecal} \\
WirbelJ\_2018 & $125$ &\citet{wirbel2019meta} \\ 
HallAB\_2017 &  $259$ &\citet{hall2017novel} \\ 
NielsenHB\_2014 & $396$ &\citet{nielsen2014systematic} \\ 
XieH\_2016 & $250$  &\citet{xie2016shotgun} \\ 
HanniganGD\_2017 & $81$ &\citet{hannigan2017viral} \\ 
Obregon-TitoAJ\_2015 & $58$ &\citet{obregon2015subsistence} \\ 
YachidaS\_2019 & $616$ &\citet{yachida2019metagenomic} \\ 
HansenLBS\_2018 & $207$ &\citet{hansen2018low} \\ 
PasolliE\_2019 & $112$ &\citet{pasolli2019extensive} \\ 
YassourM\_2016 &  $36$ & \citet{yassour2016sub}\\
YassourM\_2018 &  $271$ & \citet{yassour2018strain}\\
Heitz-BuschartA\_2016 &  $53$ & \citet{heintz2016integrated}\\
PehrssonE\_2016  &  $191$ & \citet{pehrsson2016interconnected}\\
QinJ\_2012 & $363$ & \citet{qin2012metagenome}\\
QinN\_2014 & $237$ & \citet{qin2014alterations}\\
YuJ\_2015 & $128$ & \citet{yu2017metagenomic}\\
YeZ\_2018 & $65$ & \citet{ye2018metagenomic}\\
HMP\_2012 & $748$ & \citet{human2012structure}\\
HMP\_2019\_ibdmdb & $1627$ & \citet{schirmer2018dynamics, lloyd2019multi}    \\                
HMP\_2019\_t2d & $296$ & \citet{zhou2019longitudinal}\\
RampelliS\_2015  & $38$ & \citet{rampelli2015metagenome}\\
ZeeviD\_2015 & $900$ & \citet{zeevi2015personalized}\\
IaniroG\_2022  & $165$ & \citet{ianiro2022gut}     \\
RaymondF\_2016  & $72$ & \citet{raymond2016initial}      \\
ZellerG\_2014 &  $156$ & \citet{zeller2014potential}\\
IjazUZ\_2017  & $94$ & \citet{ijaz2017distinct}      \\
RosaBA\_2018   & $24$ & \citet{rosa2018differential}  \\
ZhuF\_2020 & $171$ & \citet{zhu2020metagenome} \\
JieZ\_2017 & $385$ & \citet{jie2017gut} \\ 
\bottomrule
\end{tabular}
\label{tab:studies}
\end{table}

% \section{Discussion}

% How to identify the influence of the features? (positive or negative) 
% Limitations 

\begin{figure}
    \centering
    \includegraphics[width = 0.8\textwidth]{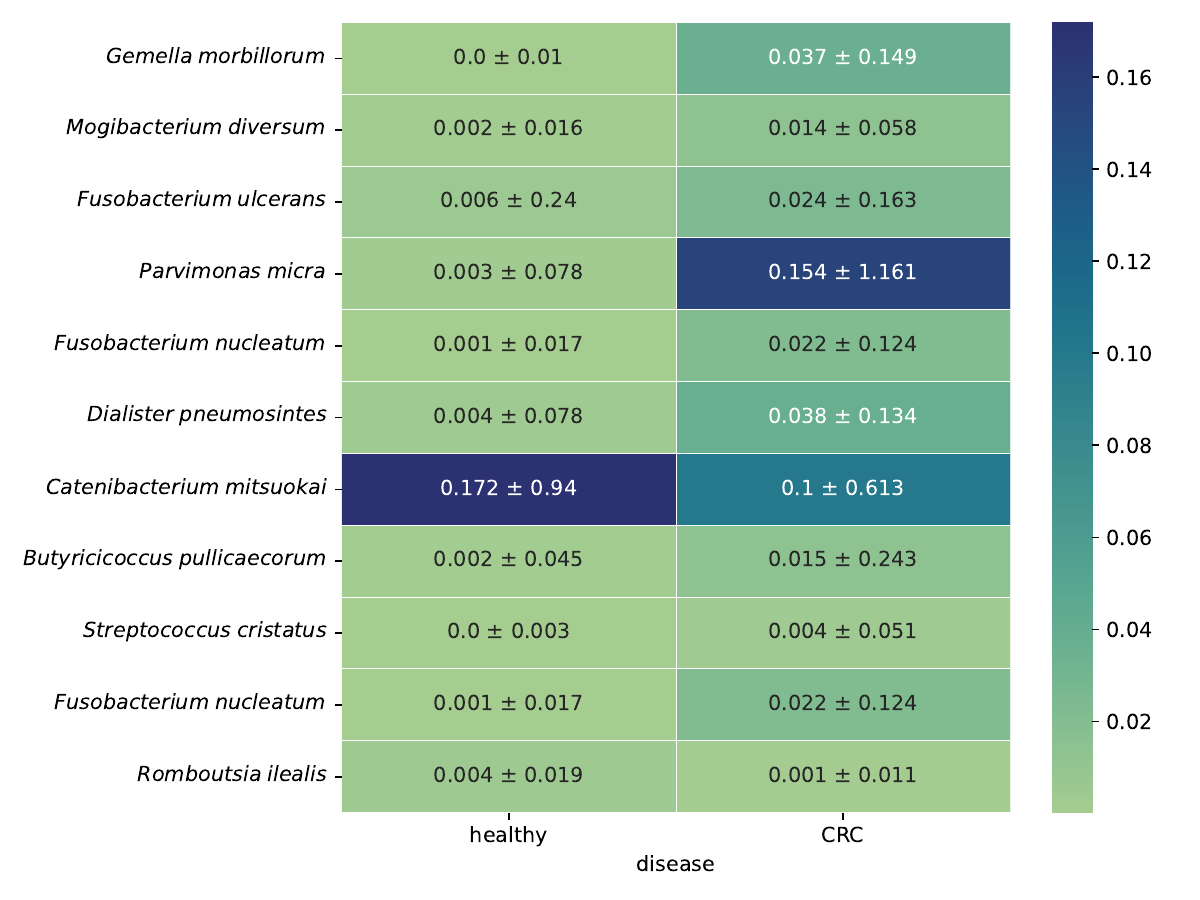}
    \caption{Mean and standard deviation values for selected species, derived from the feature importance analysis using the symbolic regression model, indicate that for CRC, most identified bacteria exhibit higher relative abundance.}
    \label{fig:mean_std}
\end{figure}

\end{document}